\documentclass[conference]{IEEEtran}
\IEEEoverridecommandlockouts
\usepackage{cite}
\usepackage{amsmath,amssymb,amsfonts}
\usepackage{algorithm}
\usepackage{algorithmic}
\usepackage{graphicx}
\usepackage{textcomp}
\usepackage{xcolor}
\usepackage{hyperref}

\def\BibTeX{{\rm B\kern-.05em{\sc i\kern-.025em b}\kern-.08em
    T\kern-.1667em\lower.7ex\hbox{E}\kern-.125emX}}

\begin{document}

\title{End-to-End Real-Time Drone-Based Person Detection Framework Using Deep Learning}

\author{
\IEEEauthorblockN{1\textsuperscript{st} Payel Sarmah}
\IEEEauthorblockA{\textit{Centre for Drone Technology} \\
\textit{Indian Institute of Technology Guwahati}\\
Guwahati, India \\
p.sarmah@iitg.ac.in}
\and
\IEEEauthorblockN{2\textsuperscript{nd} Ayush Ranjan}
\IEEEauthorblockA{\textit{Dept. of CSE} \\
\textit{KIIT University}\\
Bhubaneswar, India \\
ranjanayush881@gmail.com}
\and
\IEEEauthorblockN{3\textsuperscript{rd} Piyush Kaushik Bhattacharyya}
\IEEEauthorblockA{\textit{Dept. of CSE} \\
\textit{KIIT University}\\
Bhubaneswar, India \\
piyushbhattacharyya@gmail.com}
\and
\IEEEauthorblockN{4\textsuperscript{th} Anil Kr. Shaw}
\IEEEauthorblockA{\textit{Drone LAB} \\
\textit{National Institute of Electronics and Information Technology}\\
Bhubaneswar, India \\
anilshaw2785@gmail.com}
\and
\IEEEauthorblockN{5\textsuperscript{th} Pradip Kr. Das}
\IEEEauthorblockA{\textit{Centre for Drone Technology} \\
\textit{Indian Institute of Technology Guwahati}\\
Guwahati, India \\
pkdas@iitg.ac.in}
}

\maketitle

\begin{abstract}
In recent years, Unmanned Aerial Vehicles (UAVs) or drones have gained rapid response in terms of security, search and rescue (SAR), border surveillance, etc. Existing monitoring frameworks often struggle to maintain detection consistency when targets undergo significant scale variations due to altitude changes, leading to critical information gaps. To address this issue, this work proposes an integrated real-time detection pipeline for detecting targets through the wireless live drone video feed. Build upon YOLOv8-nano architecture, extensive flight experiments were conducted to determine the detection performance across multiple flight altitudes. Trained on VisDrone2019 dataset, the results of YOLOv8-nano model achieves 57.4\%, 41\%, 44.8\% and 20.3\% in precision, recall, mAP and mAP50:95 respectively. While demonstrating on real environment, this analysis revealed that the algorithm achieves near-total detection reliability at altitudes between 16 and 25 meters with the detection frame rate consistently maintained above 41 FPS and reaching a peak of 50 FPS. However, the goal of this work is to enable real-time person detection from an aerial platform via wireless transmission. This approach effectively addresses the dual challenges of identifying targets at varying scales and ensuring near-to-accurate localization during aerial observation.
\end{abstract}

\begin{IEEEkeywords}
Person Detection, YOLOv8, Deep Learning, Real-Time, UAV.
\end{IEEEkeywords}

\section{Introduction}

Real-time person detection from aerial platforms has become an important component of modern intelligent surveillance systems. It is increasingly used in applications such as disaster management, building monitoring, border security, large-scale crowd monitoring, etc. where rapid situational awareness is essential. Compared to ground-based sensors, Unmanned Aerial Vehicles (UAVs) offer a broader field of view and flexible deployment, allowing operators to observe complex scenes that would otherwise be difficult or unsafe to access. In emergency situations, this capability enables first responders to evaluate affected regions from a safe distance, while in public security scenarios, aerial monitoring supports efficient oversight without direct physical presence.

Traditional approaches to person detection relied heavily on manual observation or classical computer vision techniques. Methods such as background subtraction \cite{ref1}, frame differencing \cite{ref2} and Gaussian mixture models \cite{ref3} were effective in controlled settings with static cameras. Yet these techniques degrade rapidly when deployed on UAVs. The continuous motion of the drone introduces global scene changes that render background subtraction ineffective. Furthermore, aerial imagery presents unique difficulties including varying altitudes, steep viewing angles and rapid illumination changes caused by cloud cover or shadows. These factors often lead to missed detections when using non-learning-based algorithms.

The paradigm shift toward deep learning has successfully addressed many of these limitations. Convolutional Neural Networks (CNNs) have become the standard for visual perception tasks. Specifically, the YOLO (You Only Look Once) \cite{ref4} architecture revolutionized the field by treating object detection as a single regression problem rather than a complex multi-stage pipeline. Since its inception by Redmon et al. \cite{ref6}, the YOLO family has evolved through multiple iterations to balance speed and accuracy.

Recent research has focused on applying these architectures to the aerial domain. Mokayed et al. \cite{ref7} demonstrated the feasibility of deep learning in surveillance by combining YOLOv3 with DeepSORT tracking algorithms. Similarly, Krishnachaithanya et al. \cite{ref8} explored lightweight models like MobileNet-SSD to enable detection on resource-constrained devices. With the release of YOLOv8, Shyaa et al. \cite{ref9} reported significant improvements in mean Average Precision (mAP) over previous versions. Many existing studies evaluate detection accuracy on static datasets or offline video recordings. However, a study by Wang et al. \cite{ref10} proved that an improved YOLOv11-based framework can be used for UAV-assisted specifically for skier monitoring in real time.

Despite these advancements, a gap exists for sparse object detection with background complexities and detecting objects in real-time. Therefore, this work addresses these practical challenges by presenting a real-time UAV-based person detection system. Here, we propose a prototype that integrates a commercial a drone with a GPU-accelerated workstation to perform real-time inference using the pre-built YOLOv8-nano architecture. Unlike offline evaluations, this system processes live aerial footage to provide immediate visual feedback. The study specifically investigates the operational trade-offs between flight altitude, camera angle and detection performance to establish guidelines for effective aerial monitoring. The primary contributions of this paper are summarized as follows:
\begin{itemize}
\item Design and implementation of a UAV-based person detection pipeline that integrates YOLOv8n detection with wireless video streaming from a UAV platform.
\item Analyze the relationship between flight altitude and corresponding FPS, providing insights for optimal UAV deployment.
\item Develop a real-time visualization system that provides intuitive display of detection results and performance metrics.
\end{itemize}

\section{Methodology}

The proposed pipeline consists of three main stages: (1) Live video streaming from the UAV platform, (2) GPU-accelerated object detection using YOLOv8-nano model and (3) real-time performance monitoring as shown in Fig. 1. The proposed system performs real-time processing of live video footage, achieving high frame-rate person detection while maintaining stable and responsive performance.

\begin{figure}[htbp]
\centering
\includegraphics[width=0.48\textwidth]{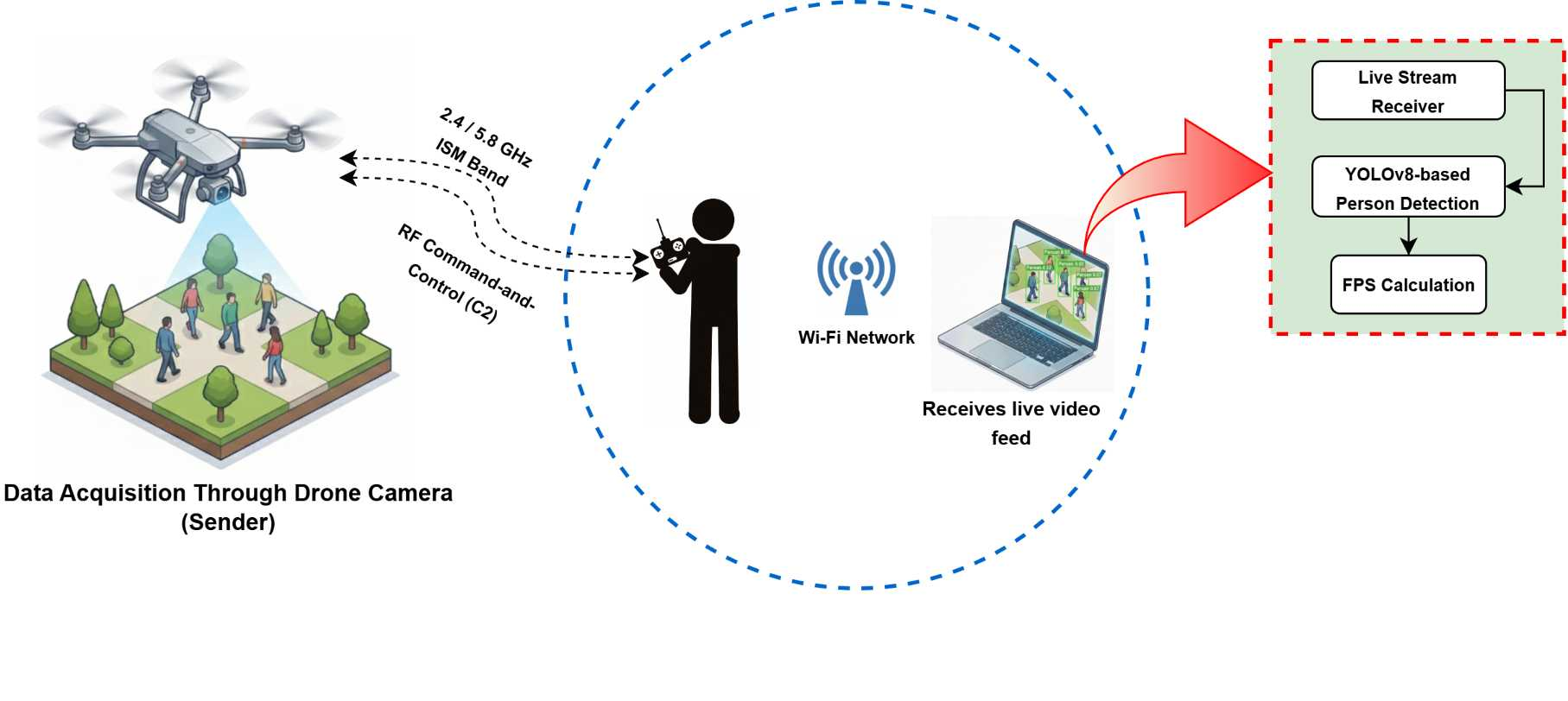}
\caption{End-to-End system architecture of the proposed framework}
\label{fig:architecture}
\end{figure}

\subsection{Live Feed Acquisition}

The foundation of the proposed system is the continuous video link established between the aerial platform and the ground processing station. Since the drone's onboard processor lacks the power for heavy deep learning inference, the system transmits the video footage to a high-performance workstation for analysis. The data flow begins with the drone camera capturing high-definition video. This footage is transmitted wirelessly to the remote controller and the connected mobile device. To bridge this data to the processing computer the mobile interface functions as a local network relay. It broadcasts the live video feed over the local network using a standard wireless transmission protocol. Thereafter, this live video feed is handled by the GPU-accelerated workstation which captures the incoming data stream and decodes it into individual image frames for immediate processing, while the drone remains free to focus on flight stability and navigation. We prioritize a low-latency local connection to ensure the detection system remains synchronized with the physical movement of the targets.

\subsection{GPU-Accelerated Detection}

To meet the low-latency requirements of aerial surveillance, the detection pipeline is optimized for strict GPU execution. The system leverages the parallel computing capabilities of CUDA-enabled hardware to accelerate matrix operations, bypassing CPU bottlenecks that typically introduce latency in real-time video processing.

The YOLOv8-nano model is employed for its optimal balance between detection accuracy and inference speed. For every input frame captured by the UAV, the model performs single-stage object detection and outputs a set of predicted bounding boxes and confidence scores in real time. Each detected object is represented by a bounding box defined by its top-left and bottom-right coordinates $(x_1^{(i)}, y_1^{(i)}, x_2^{(i)}, y_2^{(i)})$, along with a confidence score $s^{(i)}$ and a predicted class label $c^{(i)}$. Thus, in general it can be represented as $d_i$ in Eq. (1)

\begin{equation}
d_i = \left(x_1^{(i)}, y_1^{(i)}, x_2^{(i)}, y_2^{(i)}, s^{(i)}, c^{(i)}\right)
\label{eq1}
\end{equation}

To remove low-confidence and irrelevant predictions, the set of detected objects is further filtered based on confidence and class constraints. The corresponding formulation is given in Eq. (2)

\begin{equation}
\mathcal{D}' = \left\{ d_i \in \mathcal{D} \;\middle|\; s^{(i)} > \theta \wedge c^{(i)} = 0 \right\}
\label{eq2}
\end{equation}

Here, $\mathcal{D}'$ denotes the final set of valid detections obtained after post-processing. A confidence threshold $\theta$ is applied to suppress weak and irrelevant detections, such that $d_i$ with confidence scores $s^{(i)}$ greater than $\theta$ are retained. Furthermore, since this work focuses on person (pedestrian) detection, only predictions belonging to the pedestrian class ($c^{(i)} = 0$) are considered.

The adoption of the YOLOv8 architecture, particularly the Nano variant, demonstrates measurable computational benefits that are critical for this UAV-based implementation. Unlike predecessor models (e.g., YOLOv5) that rely on rigid anchor box priors, YOLOv8's anchor-free detection head eliminates the need for manual anchor tuning. This directly reduces the computational overhead during the bounding box regression phase, preventing the misalignment issues often observed when detecting small, vertically-viewed targets.

\subsection{Performance Metrics}

This study uses precision (P) recall (R) mean average precision with IoU threshold of 0.5 (mAP50), and average precision with IoU thresholds varying from 0.5 to 0.95 (mAP50:95) as evaluation metrices for the model. Precision refers the proportion of the dataset that are detected as targets are actually targets. Recall calculates the proportion of the actual targets are detected successfully by the model. The formulae for calculating precision and recall are shown in Eq. 3:

\begin{equation}
\text{Precision} = \frac{TP}{TP + FP}, \quad \text{Recall} = \frac{TP}{TP + FN}
\label{eq3}
\end{equation}

Where TP, FP, and FN refers the number of pedestrians accurately detected, the number of non-pedestrians are mistakenly detected as pedestrians, and FN refers to when the model fails to detect a pedestrian who is actually present respectively.

Generally, a higher mean average precision (mAP) indicates stronger overall model performance. The formulae used to calculate mAP50 and mAP50:95 are provided in Eq. 4 and 5 respectively:

\begin{equation}
\text{mAP@50} = \frac{1}{N} \sum_{N_C=1}^{N} AP_C
\label{eq4}
\end{equation}

\begin{equation}
\text{mAP@50:95} = \frac{1}{k} \sum_{IoU=0.5}^{0.95} mAP_{IoU}
\label{eq5}
\end{equation}

where $N_C$ is the number of classes and $k$ is the number of times the mAP is calculated individually for a single IoU threshold.

\subsection{Real-Time Performance Monitoring}

In UAV-based surveillance, the system's responsiveness is as critical as its detection accuracy. To quantify the temporal efficiency of the proposed pipeline, the system employs a continuous performance monitoring mechanism that calculates the processing latency for every individual video frame. The total processing time per frame, denoted as $\tau_{proc}$ encompasses the duration required for frame acquisition, pre-processing such as resizing and normalization, GPU inference and post-processing. This is measured by capturing high-resolution timestamps at the initiation $t_{start}$ and conclusion $t_{end}$ of the detection loop. The processing latency for the $k$-th frame is defined in Eq. (6)

\begin{equation}
\tau_{proc}^{(k)} = t_{end}^{(k)} - t_{start}^{(k)}
\label{eq6}
\end{equation}

Subsequently, the instantaneous Frame Rate (FPS), which represents the system's throughput, is derived as the reciprocal of the processing latency. To ensure numerical stability and prevent potential division-by-zero errors in cases of negligible processing time, a small epsilon term ($\epsilon \approx 10^{-6}$) is added to the denominator. The instantaneous frame rate $f_k$ is calculated as shown in Eq. (7):

\begin{equation}
f_k = \frac{1}{\tau_{proc}^{(k)} + \epsilon}
\label{eq7}
\end{equation}

The metric $f_k$ is computed dynamically during flight and overlayed directly onto the video feed. This allows for immediate visual verification of the system's speed, ensuring that the inference rate remains synchronized with the RTMP stream's input rate, thereby preventing buffer overflow and minimizing video lag.

\subsection{Detection Visualization}

The core workflow of the proposed system is summarized in Algorithm 1. This algorithm outlines the complete pipeline, starting from the initialization of hardware resources to the final visualization of detection results. It takes the live video stream and the YOLOv8 model as inputs and generates an annotated live video stream with bounding boxes and performance metrics (FPS) in real time.

\begin{algorithm}
\caption{Real-Time UAV Person Detection Pipeline}
\begin{algorithmic}[1]
\STATE \textbf{INPUT:} Video Stream $S$, Threshold $\theta$, Model $M$
\STATE \textbf{OUTPUT:} Annotated Frame $F_{out}$ with bounding box
\STATE \textbf{INITIALIZE:} Verify CUDA availability; HALT if Null
\STATE $M \leftarrow$ LoadModel("yolov8n.pt") on GPU device
\STATE $V \leftarrow$ Connect($S$)
\WHILE{$V$ is Active}
    \STATE $t_{start} \leftarrow$ CurrentTime()
    \STATE $F_{in} \leftarrow$ ReadFrame($V$)
    \IF{$F_{in}$ is Null}
        \STATE Reconnect to $S$
        \STATE Continue
    \ENDIF
    \STATE $D_{raw} \leftarrow$ Predict($M$, $F_{in}$)
    \FOR{each detection $d_i \in D_{raw}$}
        \STATE Extract $(x_1, y_1, x_2, y_2, s, c)$ from $d_i$
        \IF{$s > \theta$ AND $c = 0$}
            \STATE DrawBox($F_{in}$, $x_1$, $y_1$, $x_2$, $y_2$)
            \STATE AddLabel($F_{in}$, $s$)
        \ENDIF
    \ENDFOR
    \STATE $t_{end} \leftarrow$ CurrentTime()
    \STATE $FPS \leftarrow \frac{1}{(t_{end} - t_{start}) + \epsilon}$
    \STATE Overlay($F_{in}$, $FPS$)
    \STATE Display($F_{in}$)
\ENDWHILE
\end{algorithmic}
\end{algorithm}

\section{Implementation and Results}

The system was tested with live aerial footage from the drone at multiple altitudes. Various performance metrics has been implemented to evaluate the model.

\subsection{Dataset Description}

To ensure the model remains robust against common aerial photography challenges such as high-angle viewpoints and varying object scales the YOLOv8n architecture was fine-tuned using the VisDrone2019 dataset \cite{ref5} dataset.

The dataset comprises 10,209 static images and 263 video sequences captured across diverse scenarios including urban landscapes and rural terrain under fluctuating lighting conditions. It features a broad range of resolutions with static images reaching 2000$\times$1500 pixels and video sequences recorded at up to 4K resolution. To align with the input tensor constraints of the YOLOv8n architecture all input imagery was resized to a standard 640$\times$640 resolution. The annotations were preprocessed to prioritize human surveillance objectives by merging the distinct 'Pedestrian' and 'People' categories into a single unified 'Person' class. Consequently the dataset was partitioned into 6,471 training images, 548 validation images and 1,610 testing images to rigorously evaluate the model's generalization capabilities.

\subsection{Experiment Details}

This study employs the YOLOv8-nano model for the detection tasks. A comprehensive list of the hardware specifications and hyperparameters used during the training phase is provided in Table I. To evaluate the geometric alignment of detections, we utilize the Intersection over Union (IoU) metric, representing the area of intersection divided by the area of union between the predicted and actual boxes.

\begin{table}[htbp]
\caption{Experimental Configuration and Training Hyperparameters}
\label{tab:hyperparams}
\centering
\begin{tabular}{ll}
\hline
\multicolumn{2}{c}{\textit{System Environment}} \\
\hline
\textbf{Parameter} & \textbf{Configuration} \\
\hline
Operating System & Windows 11 \\
Programming Language & Python 3.10.19 \\
Deep Learning Framework & PyTorch (with Ultralytics YOLO) \\
Processor (CPU) & Core(TM) i7-13620H \\
Graphics Unit (GPU) & NVIDIA GeForce RTX 4060 \\
Acceleration & CUDA 13 \\
\hline
\multicolumn{2}{c}{\textit{Training Hyperparameters}} \\
\hline
Model Architecture & YOLOv8-Nano (yolov8n.pt) \\
Input Resolution & 640 $\times$ 640 pixels \\
Batch Size & 64 \\
Total Epochs & 200 \\
Optimizer & SGD \\
Learning Rate ($lr$) & 0.01 \\
IoU Threshold & 0.5 \\
\hline
\end{tabular}
\end{table}

\subsection{Performance Result}

To comprehensively evaluate the proposed UAV-based detection framework, we assessed the model's performance on the VisDrone2019 test set using both quantitative metrics and visual inspection of real-world inference scenarios. The evaluation focuses on the YOLOv8-Nano model's ability to balance detection accuracy with the constraints of aerial imagery, such as small object scales and high-density crowds. The peak quantitative performance metrics achieved after the 200-epoch training cycle are summarized in Table II. As observed, the model attained a precision of 0.574 and a recall of 0.410. This distribution is characteristic of aerial surveillance tasks; the higher precision indicates that the system is highly reliable when it flags a detection, effectively filtering out non-human background clutter like trees or street poles. The mAP@0.5 score of 0.448 further validates that despite the lightweight nature of the nano architecture, it retains sufficient feature extraction complexity to handle the diverse environments found in the VisDrone benchmark. Furthermore, the mAP@0.5:0.95 score of 0.203 provides a stricter assessment of localization quality. This metric, which averages performance across increasing IoU thresholds (from 0.5 to 0.95), confirms that the model does not merely detect the presence of a person but generates bounding boxes that fit the target boundaries with a reasonable degree of pixel-level accuracy, despite the challenge of delineating small objects from high altitudes.

\begin{table}[htbp]
\caption{Quantitative Performance}
\label{tab:quant}
\centering
\begin{tabular}{lc}
\hline
\textbf{Performance Metric} & \textbf{Achieved Value} \\
\hline
Precision & 0.574 \\
Recall & 0.410 \\
mAP (IoU=0.5) & 0.448 \\
mAP50:95 & 0.203 \\
\hline
\end{tabular}
\end{table}

To understand the learning dynamics that led to these results, we analyzed the training history illustrated in Fig. 2. The curves for precision, recall and mAP exhibit a distinct logarithmic growth pattern. During the initial 50 epochs, the model demonstrates rapid feature learning, quickly adapting to the vertical perspective of the UAV imagery. Following this phase, the optimization stabilizes, with the metrics plateauing between epoch 150 and 200.

\begin{figure}[htbp]
\centering
\includegraphics[width=0.48\textwidth]{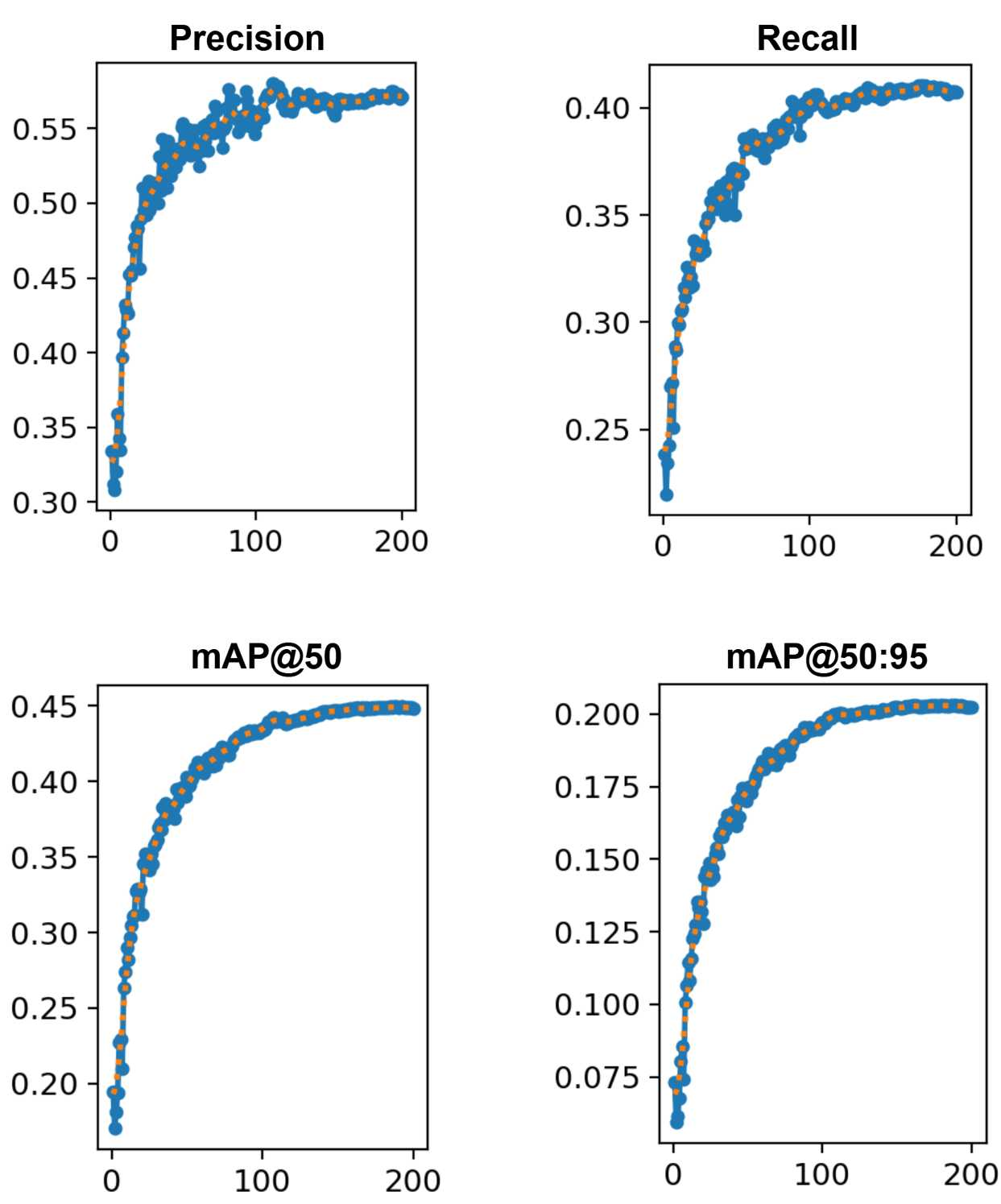}
\caption{Training progression of detection metrics over 200 epochs. The curves for Precision, Recall, mAP@50 and mAP@50:95 indicates a stable learning trajectory, with rapid convergence observed in the first 50 epochs followed by a consistent plateau, demonstrating robust generalization.}
\label{fig:training}
\end{figure}

To verify the robustness and generalization capability of the trained model, we performed a visual inspection of the inference results on the test dataset. Fig. 3 presents the inference results across three distinct environments, chosen to represent the core challenges of aerial surveillance: crowd density, illumination variation and small-scale targets. Fig. 3(a) depicts a high-density urban plaza characterized by significant inter-object occlusion. In this scenario, multiple pedestrians are moving in close proximity which creates complex overlap patterns. The model demonstrates a high retrieval rate and successfully identifies the majority of targets within the crowd. However, a limitation is observed in this scenario. Due to the significant inter-object occlusion and the proximity of targets, the system occasionally generates redundant overlapping bounding boxes for the same individual. However, the performance of the model under complex illumination conditions is evaluated in Fig. 3(b) which features a busy street illuminated by bright signs and lights. Such environments often introduce background noise and glare that can confuse optical sensors. However, the model maintains robust detection precision by accurately identifying pedestrians despite low ambient light and visual clutter which indicates that the trained model has learned feature representations that are largely invariant to drastic changes in lighting distribution. Here, in Fig. 3(c) show a broad view of a city area taken from high-altitude. In this view, the effective spatial resolution of the targets is drastically reduced, with human targets occupying minimal pixel area relative to the frame size. The successful localization of these tiny objects validates the effectiveness of the multi-scale feature fusion within the YOLOv8 architecture, confirming its ability to retain semantic information even when targets undergo significant down-sampling due to altitude.

\begin{figure}[htbp]
\centering
\includegraphics[width=0.48\textwidth]{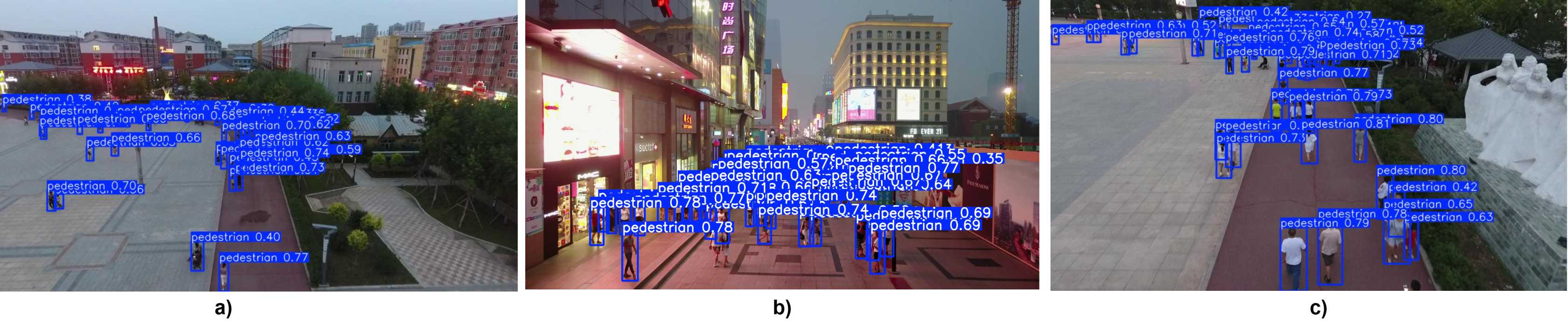}
\caption{Qualitative inference results on the VisDrone2019 test set. (a) Detection in a crowded urban plaza (b) Robust performance in a low-light commercial district (c) Small-object detection from a high-altitude and wide-angle aerial perspective.}
\label{fig:qualitative}
\end{figure}

\section{On-Site Experimental Validation}

To validate the practical viability of the proposed YOLOv8-Nano framework, an extensive field trial was conducted at the Indian Institute of Technology (IIT) Guwahati which can be found here: \url{https://youtu.be/4Kj43RIqvwc?si=p1Zl0UNsifCPsuIK}. The flight tests were performed in an outdoor campus environment featuring mid-complex background textures (grass, pavement, and architectural structures) and dynamic pedestrian movement. The aerial footage was acquired using a DJI Mini 4 Pro drone maneuvering at altitudes ranging from 15 to 30 meters, with camera gimbal pitches adjusted between $-35^{\circ}$ and $-90^{\circ}$. The live video feed was processed in real-time using the proposed framework.

To provide a granular analysis of the system's performance during the field test at IIT Guwahati, we logged real-time telemetry synchronized with the inference metrics. Table III correlates the flight parameters, specifically, altitude, camera gimbal pitch, and detection stability. The real-time detection results can be found in Fig. 4. The empirical data reveals a distinct operational optimum for the YOLOv8-Nano model within the UAV environment. Specifically, during the flight phase at an altitude of 25 meters with a camera gimbal pitch of approximately $-50^{\circ}$ with 48.53 FPS, the system exhibited its peak performance as shown in Fig. 4(b). In these regions, the high contrast between the subjects (dark clothing) and the uniform concrete surface allows the YOLOv8 feature extractor to easily delineate bounding boxes, even though the drone is flying quite high. On the right side, there are banners, tables, and signs and this busy background confuses the model because the colors and shapes of the banners look somewhat similar to the people standing nearby, causing the drone to miss some targets in that specific spot. Also, because the camera is tilted down steeply at $-50^{\circ}$, the people look shorter and more squashed than they would from a normal eye-level view. While the person seen in the foreground grass, target located further back in the grassy areas occasionally missed as shown in Fig. 4(a). This basically attributes to the chromatic homogeneity where the spectral properties of the green landscaping blend with the subjects, particularly if they are wearing low-contrast clothing.

\begin{table}[htbp]
\caption{Field Performance Log: IIT Guwahati Flight Test}
\label{tab:field}
\centering
\begin{tabular}{cccl}
\hline
\textbf{Altitude} & \textbf{Angle ($\theta$)} & \textbf{Avg FPS} & \textbf{Qualitative Assessment} \\
\hline
(a) 16 m & $-50^{\circ}$ & 35.31 & Proximal Target Separation \\
(b) 20 m & $-35^{\circ}$ & 41.11 & Complex Scene Localization \\
(c) 25 m & $-50^{\circ}$ & 48.53 & Motion-Robust Tracking \\
\hline
\end{tabular}
\end{table}

At the altitude of 20 meters and $-35^{\circ}$ camera angle with the FPS of 41.11, the people walking on the paved road stand out very clearly against the background by recognizing and drawing accurate bounding box around them. However, the system struggles a bit in more cluttered parts of the image. Furthermore, when the drone was maneuvered to a lower altitude of 16 meters with a camera angle of $-50^{\circ}$ degrees with 35.31 FPS, as shown in Fig. 4(c), it was found that the detection stability improved significantly compared to the higher altitude flights. In this specific frame, the system accurately localized two stationary individuals standing in the lawn. The main reason behind this is the spatial resolution obtained by descending the drone to 16 meters and at this closer range, the targets occupy a larger portion of the frame, allowing the model to extract much richer detail from the human figures against the green grass background by creating a strong contrast that makes the targets easier for the neural network to recognize. Moreover, object detection indicates that convolutional networks perform best when the foreground object possesses a distinct intensity gradient compared to the background. Here, the targets who are wearing white and black created sharp and high-frequency edges against the low-frequency, uniform texture of the grass. However, due to the static appearance of the targets, the video feed was free from motion blur that often degrades performance during rapid movements which allowed the camera to capture sharp and well-defined edges around the targets, ensuring that the model could lock onto them without ambiguity.

\begin{figure}[htbp]
\centering
\includegraphics[width=0.48\textwidth]{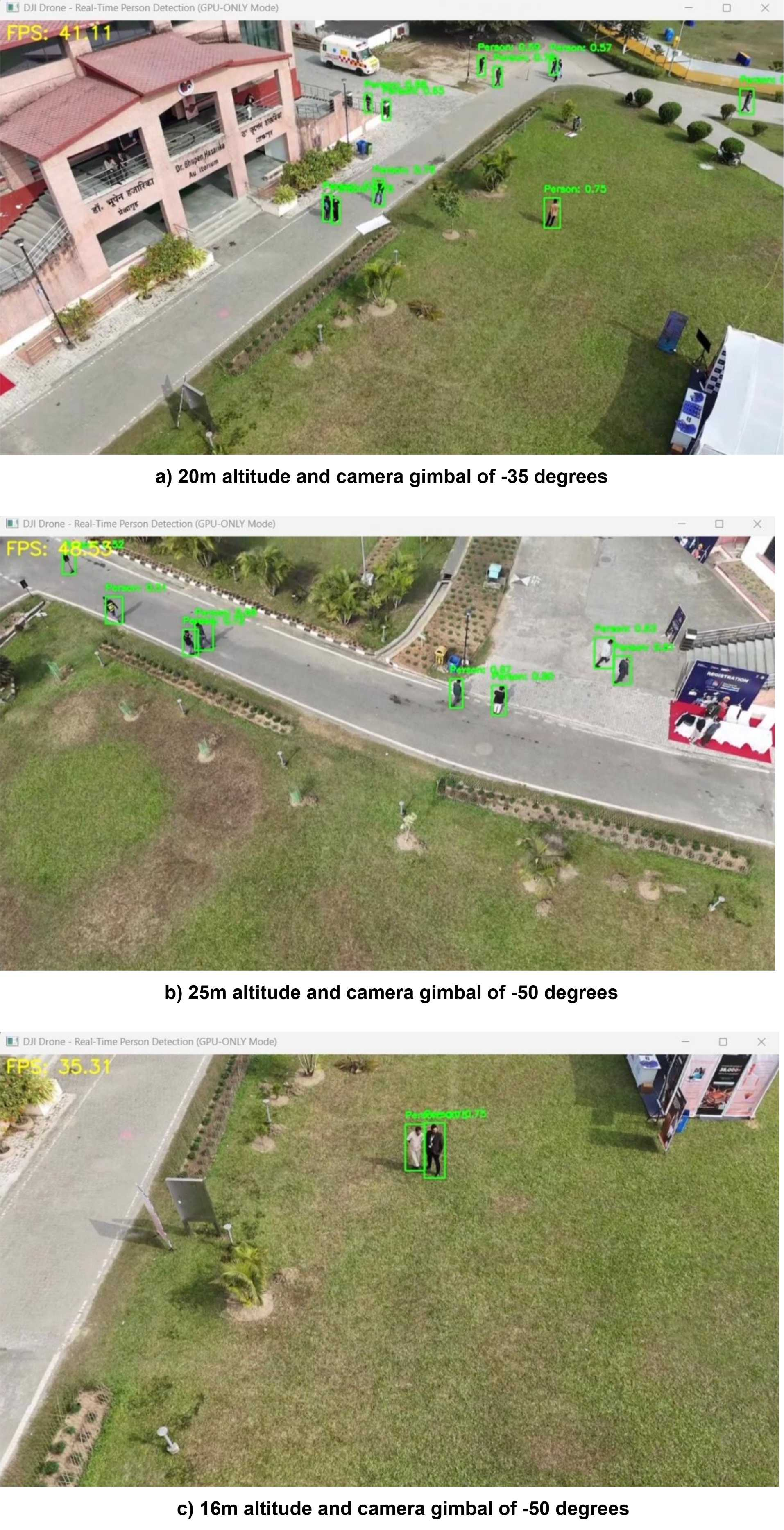}
\caption{Real-time detection results illustrating the impact of flight geometry on model performance. (a) Balanced coverage at 20m (b) Achieved peak system performance of 48.53 FPS with motion-robust tracking (c) High-resolution feature extraction at 16m altitude.}
\label{fig:fieldtest}
\end{figure}

\section{Conclusion}

This paper presented a comprehensive design and implementation of a real-time, UAV-based person detection system that combines the strengths of YOLOv8n-based object detection with wireless video streaming from a drone platform. Field validation at the real world environment confirmed that the model fine-tuned on VisDrone2019 dataset adapts well to real outdoor environments. The study also identified specific boundaries where detection degrades. Reliability diminishes when targets appear at the extreme edge of the live video stream or from a strictly vertical top-down perspective due to the lack of visible features. Future work will address these constraints by incorporating temporal tracking to maintain identity across frames. We also aim to migrate processing to onboard edge units to enable fully autonomous operations in areas without communication.

\section*{Acknowledgment}

The authors would like to acknowledge that this work is a collaborative project supported by the SwaYaan project at the National Institute of Electronics Information Technology (NIELIT), Bhubaneswar. This research was sponsored by the Ministry of Electronics and Information Technology (MeitY), Government of India.

\end{document}